\documentclass[11pt]{article}

\usepackage[preprint]{acl}
\usepackage{amsmath}
\usepackage{longtable}
\usepackage{times}
\usepackage{latexsym}
\usepackage{tabularx}
\usepackage{booktabs}
\usepackage{array}

\newcolumntype{Y}{>{\raggedright\arraybackslash}X}
\usepackage[T1]{fontenc}

\usepackage[utf8]{inputenc}

\usepackage{microtype}

\usepackage{inconsolata}

\usepackage{graphicx}
\usepackage{booktabs}
\title{Error-Supervised Synthetic Learner Writing for Automated Essay Scoring}

\author{
\textbf{Duy Anh Nguyen\textsuperscript{1,2}} \\
\textsuperscript{1}University of Greenwich, London, United Kingdom \\
\textsuperscript{2}FPT University, Can Tho, Vietnam \\
\texttt{na6367q@gre.ac.uk} \\
\texttt{anhndgcc240003@gmail.com}
}

\begin{document}
\maketitle
\begin{abstract}
Synthetic essays can help reduce dependence on human-written data in Automated Essay Scoring (AES). However, they often lack realistic errors, limiting their ability to represent authentic human writing, particularly when the target texts are intended to resemble those produced by language learners. In this study, we present a simple approach that introduces error supervision into synthetic essay generation. Specifically, we fine-tune an LLM generator on error-annotated texts of the kind commonly used in Grammatical Error Detection (GED). To assess the utility of the proposed approach, we fine-tune and evaluate AES scorers under three data conditions: authentic essays, synthetic essays generated conventionally, and synthetic essays generated using our proposed approach. The results show that in the larger-data settings, the proposed approach outperforms the conventional synthetic baseline in 11 out of 12 dataset-metric comparisons, with performance in some cases approaching that of models trained on authentic essays. Despite these gains, performance under extremely low-resource settings remains mixed, with advantages over the conventional baseline only becoming more apparent at 200 training essays, although not consistently across datasets. Qualitative and quantitative analyses further show that the proposed approach produces learner-like errors whose distributions broadly resemble those observed in authentic essays.
\end{abstract}

\section{Introduction}
Synthetic data has been widely used as a means of addressing low-resource settings in which sufficient amounts of authentic data may be unavailable or difficult to obtain. This is particularly relevant to Automated Essay Scoring (AES), where collecting large-scale datasets of human-written essays with reliable score annotations can be costly and time-consuming. Synthetic essay generation therefore offers a potential way to supplement limited authentic training data and reduce dependence on large quantities of manually scored essays.

However, a persistent challenge in synthetic essay generation is that generated texts may be overly well-formed and linguistically accurate \cite{yoo-etal-2025-dress, ETAAT2026100258}, making them less representative of authentic human writing. This is especially problematic in AES for English learner writing, where writers are still developing in terms of language proficiency and naturally produce a range of grammatical and lexical errors. If synthetic essays fail to reproduce these proficiency-related error patterns, they may not adequately reflect the characteristics of the learner essays that AES systems are ultimately expected to evaluate.

At the same time, several learner corpora containing explicit error annotations have been developed and extensively used in research on Grammatical Error Detection (GED) and Grammatical Error Correction (GEC) \cite{bryant-etal-2023-grammatical}. However, these annotations have primarily been developed and studied in the context of GED and GEC. Such corpora provide detailed information about the types and distributions of errors produced by language learners, creating an opportunity to use them beyond error detection and correction. In particular, these annotations could be incorporated into synthetic essay generation to encourage models to produce learner essays with more realistic and proficiency-aligned error patterns, potentially resulting in synthetic data that better represent authentic learner writing.

In this study, we propose a simple approach to improve the utility of synthetic learner essays for downstream AES by fine-tuning essay generators on error-annotated learner writing typically used in GED and GEC research. We evaluate the approach across three learner corpora by comparing AES scorers trained on the proposed synthetic data with those trained on authentic data and a conventional synthetic baseline, including one corpus whose error annotations are derived automatically using GEC and GED tools. We further conduct qualitative and quantitative analyses to assess how closely the generated essays resemble authentic learner writing. Together, these experiments address the following research questions:
\begin{enumerate}
    \item{\textbf{RQ1}} Does synthetic learner writing generated using error-tag supervision provide greater benefit for downstream AES than synthetic writing generated directly from unannotated essays?
    \item{\textbf{RQ2}} To what extent do the error patterns produced by the proposed approach resemble those found in authentic learner writing at the corresponding proficiency level or score?
\end{enumerate}
\section{Related Work}

\subsection{Synthetic Data Augmentation for AES}
AES systems depend heavily on the availability of score-annotated essays. However, collecting sufficiently large datasets of human-scored writing is often time-consuming, costly, and resource-intensive. To address this limitation, a growing number of researchers have explored the use of synthetic data to expand AES training datasets. An early study by \citet{Park-2022-essayGAN} introduced a method for generating score-conditioned synthetic essays and adding them to the training data of AES models. More recently, \citet{Zhang-et-al-2026} used DeepSeek-V3 with score-specific demonstrations to generate 1,200 essays. They subsequently trained an AES model on a combination of authentic and synthetic data and reported improvements on the ASAP dataset. A separate recent study investigated the use of GPT-4 and GPT-4o for AES augmentation, finding that synthetic responses could provide training utility comparable to authentic student responses under some experimental settings, despite the performance varying across prompts and training sample sizes \citep{Zhang_Badola_Johnson_Li_2026}.

\subsection{Controlled Synthetic Essay Generation}
Artificial error generation has been extensively studied in GED and GEC, where errors are introduced into otherwise well-formed text to create additional training data \cite{rei-etal-2017-artificial, bryant-etal-2023-grammatical}. In the context of AES, a prominent example of synthetic essay construction through controlled corruption is \citet{yoo-etal-2025-dress}, who introduced the DREsS dataset, which includes a subset of synthetically corrupted essays known as DREsSCASE. This subset was created by applying sentence-level corruptions to high-scoring essays to produce lower-scoring versions. In contrast with the present study, DREsS introduces errors through rule-based sentence corruption rather than training a generator to learn naturally occurring, proficiency-dependent error patterns.

Although their study did not focus on English writing, \citet{qwaider-etal-2025-enhancing} introduced a highly relevant synthetic data augmentation framework based on controlled error injection. The authors prompted GPT-4o to generate 3,040 Arabic essays across different CEFR proficiency levels and subsequently introduced learner-like errors using either GPT-4o or a controlled transformer-based method. Their findings suggest that controlled error injection can outperform the use of error-free GPT-generated essays, highlighting the potential of proficiency-aware error modelling to produce more effective synthetic data for AES.

In contrast to post-generation corruption approaches introduced in the above studies, \citet{nam-2027-prompt} optimized LLM prompts to directly generate synthetic essays whose handcrafted feature distributions resembled those of authentic essays. In some experimental settings, adding these synthetic essays to a small authentic training set produced AES performance close to that achieved using the complete authentic dataset. A study by \citet{chen-etal-2026-cpt} similarly fine-tuned an instruction-tuned language model on proficiency-rated student essays to generate essays given a writing prompt and a student proficiency profile. However, unlike our proposed approach, their framework generates unannotated essays using separate models for low, medium, and high proficiency writers, which does not explicitly focus on learner errors.

Taken together, prior work either introduces learner-like errors through post-generation corruption or generates proficiency-controlled essays from unannotated learner texts. The former separates error creation from essay generation, while the latter provides no explicit supervision over learner error patterns. It therefore remains unclear whether directly supervising a generator with learner-error annotations during fine-tuning can improve the downstream AES utility of synthetic learner writing. The present study investigates this question by comparing error-supervised generation with generation from unannotated essays.

\section{Data and Preprocessing} 
We use three English learner writing corpora independently for synthetic data generation. Two provide explicit error annotations, while for the third we derive annotations automatically using GEC and GED tools. The datasets are described below.

\paragraph{CLC FCE (v1.1)}
The CLC FCE Dataset \cite{yannakoudakis-etal-2011-new} comprises 1,244 anonymized examination scripts produced by learners taking the Cambridge First Certificate in English examination. Each script contains responses to two different writing tasks. We treat each task response as a separate instance, yielding up to 2,488 texts in total across 31 prompts. The corpus additionally provides the corresponding writing prompts, assessment scores, manually annotated learner errors, and selected demographic information. The learner texts are manually error-annotated using the native Cambridge Learner Corpus taxonomy, which comprises 77 fine-grained error types. The annotations are embedded within the texts using XML elements that identify the original erroneous form, its proposed correction, and the corresponding error category. The scores assigned to each text in this dataset range from 0 to 20.

\paragraph{Write \& Improve}
The Write \& Improve Corpus \cite{nicholls-etal-2024-write-improve} contains 5,050 user--prompt sets produced by 766 English learners in response to 50 writing prompts. Each set comprises multiple versions submitted by the same learner for the same prompt as they revised their writing in response to automated scoring and feedback. The corpus contains 23,216 submitted versions in total. In this study, we use the final version of each user- prompt set, yielding 5,050 texts, 4,546 of which are error-annotated. These final versions are accompanied by human-assigned CEFR labels ranging from A1 to C2 and human-corrected reference texts produced according to a minimal-edit principle. This dataset also provides corresponding edit spans and error categories in M2 format, which were previously derived from the original--corrected pairs using ERRANT \cite{bryant-etal-2017-automatic} -- an automatic error annotation toolkit.

\paragraph{ELLIPSE}
The English Language Learner Insight, Proficiency and Skills Evaluation (ELLIPSE) Corpus \cite{ellipse} comprises 6,482 essays written by English language learners in response to 29 writing prompts. Each essay is scored by trained human raters for overall language proficiency, as well as six analytic dimensions covering cohesion, syntax, vocabulary, phraseology, grammar, and conventions. In this study, however, we only use the holistic proficiency score, which ranges from 1.0 to 5.0 in 0.5-point increments, as the target label. Since ELLIPSE does not provide explicit error annotations or corrected versions of the essays, we apply GECToR-2024 \cite{omelianchuk-etal-2024-pillars} to each original essay to obtain an automatically corrected version, and then apply ERRANT to the original--corrected pair to identify and classify the edits and construct the error-tagged targets used for model fine-tuning.

During preprocessing, all datasets are standardized into a common CSV format. Duplicate entries, empty texts, and instances with invalid or missing scores are removed. The original datasets use different ratios for train, development, and test splits. We therefore recombine each dataset and re-split it using a consistent 70/15/15 ratio. This standardization ensures that all datasets are evaluated under comparable experimental conditions. For the CLC FCE and W\&I corpora, splitting is performed at the author level so that responses written by the same learner are assigned to the same partition, thereby preventing author-level data leakage across the train, development, and test sets. For ELLIPSE, proportional stratified sampling is used to preserve the distribution of holistic proficiency scores across the three partitions. For the W\&I corpus, intermediate labels (e.g., A2+, B1+, B2+) are collapsed into their corresponding main CEFR levels. This normalization aligns the corpus with the standard CEFR proficiency categories commonly used in learner-language datasets and reduces dependence on the corpus-specific granularity.

\section{Methodology}\label{sec:methodology}
In this section, we present the proposed methodology for generating synthetic learner essays with explicit error supervision.
\subsection{Generation Task Formulation}
\label{sec:generation-task}
We formulate synthetic learner essay generation as a proficiency-conditioned sequence generation task. Given a writing prompt $P$ and a target proficiency level $L$, the model generates a complete error-tagged learner essay $T$. The tagged output is subsequently converted into an erroneous surface essay $E$ through a deterministic transformation:

\begin{equation}
    (P,L) \rightarrow T \rightarrow E.
\end{equation}

Here, $P$ represents the writing prompt or task instruction, while $L$ represents the label indicating the essay's quality, such as a single holistic essay score or the author's CEFR level. The intermediate output $T$ contains both the essay content and explicit annotations of the learner errors occurring within the essay. The final output $E$ is the corresponding surface essay in which the annotation markup has been removed while the erroneous forms have been retained. 

\subsection{Model Training and Data Generation}\label{sec:model_training}
In this study, we use Qwen2-7B-Instruct \cite{yang2024qwen2technicalreport} as the generator, as it has previously been used for the generation of synthetic essays in a highly similar generation task \cite{chen-etal-2026-cpt}, providing an established precedent for its use in this setting. Additionally, its open-weight availability makes it a practical, cost-effective choice for repeated fine-tuning and generation experiments at the scale required in this study.

We fine-tuned Qwen2-7B-Instruct with the writing prompt $P$ and target label $L$ as input, and the corresponding authentic learner essay with inline error annotations as the target output. To ensure a consistent generation target across corpora, we standardized all annotations to the format \texttt{<ERR type="ERROR\_TYPE" cor="CORRECTION">erroneous text</ERR>}, using \texttt{<MISSING/>} for missing-item errors and an empty correction attribute for unnecessary items. The original error taxonomy of each corpus was retained; only the annotation format was standardized. At inference time, the model generates essays in this annotated format. We then deterministically derive the corresponding erroneous surface essays by removing the inline error annotations using simple RegEx processing.

To make this process computationally efficient, we employed Quantized Low-Rank Adaptation (QLoRA; \citealp{QLoRA}), which enables parameter-efficient fine-tuning with substantially lower memory requirements than full-model fine-tuning. QLoRA was particularly suitable for this study because the experimental design spans three datasets and requires multiple generator training runs across different conditions. By quantizing the frozen base model to 4-bit precision, QLoRA substantially reduces GPU memory usage and training cost compared with the conventional LoRA \cite{LoRA}, while retaining the representational capacity of the pretrained model. We train a separate generator for each dataset using its corresponding training split.

At inference, synthetic essays are generated by conditioning the generators on the prompt and label associated with each essay in the corresponding authentic training set. Further details of the training and generation procedures are provided in Appendix \ref{sec:gen_details}.

\section{Evaluation of Generated Essays}\label{sec:analyses}
Before evaluating downstream AES performance, we examine the essays generated using the proposed approach to characterize their error patterns and assess how closely they resemble those of the authentic learner essays, thereby addressing \textbf{RQ2}. 

We begin by randomly sampling 30 essays from each generated dataset using proportional stratified sampling. Manual inspection revealed that the generators generally produced annotations in the specified tag format, with many errors appearing learner-like, including tense inconsistencies, missing determiners, spelling errors, and preposition errors. Error supervision also did not force errors into every text: seven sampled W\&I essays and two sampled CLC FCE essays contained no error tags. The essays generally followed the expected task and genre conventions, although occasional task or genre drift was observed. Other recurring weaknesses included over-correction, semantic inconsistencies, repetition, particularly in ELLIPSE, and occasional incorrect attribution of error types. The latter does not directly affect downstream AES because the tags are removed before scoring, but it introduces noise into the generated annotations.

To provide quantitative evidence, we compare the synthetic dataset with the authentic dataset in terms of error density, measured as errors per 100 words, and compute Jensen--Shannon Divergence (JSD; \citealp{JSD}) between their error-type distributions, with lower JSD indicating greater similarity. Both metrics are computed at each proficiency level or score. Across all three datasets, proposed synthetic essays generally exhibit lower error density than authentic essays, although the gap tends to narrow at higher proficiency levels. This is particularly pronounced in W\&I, where the two conditions are nearly identical at C2. JSD values remain relatively low overall, with most below 0.25. Specifically for the CLC FCE corpus, JSD values are predominantly below 0.10. Higher divergence occurs mainly at the extremes of the proficiency or score ranges, which also contain fewer samples and may therefore be more sensitive to sampling variability. Further details of the above analyses are provided in Appendix~\ref{sec:detailed_analyses}, including example generated essays, visualizations of the quantitative metrics, and the largest discrepancies in error-type proportions between the authentic and synthetic data.

Overall, the analyses suggest that the proposed approach generally produces essays with learner-like errors whose distributions and proficiency-related patterns broadly resemble those of the authentic training data, despite lower error density and several remaining generation and annotation weaknesses. These results provide sufficient support to proceed with evaluating the downstream utility of the generated essays for AES.

\section{AES Evaluation}
\subsection{Experimental Setup} \label{sec:experimental_setup}
To directly address \textbf{RQ1}, we compare the performance of AES models trained on three separate data conditions for each learner corpus:
\paragraph {Authentic} Original authentic essays from the training split of each dataset. These are the same learner essays used to fine-tune the corresponding generators described in Section \ref{sec:methodology}.
\paragraph{Synthetic (Proposed)} Synthetic learner essays generated using our proposed approach. The corresponding generators are trained using the error-tag supervision described in Section \ref{sec:methodology}. During generation, the resulting error tags are deterministically removed to obtain erroneous surface essays for AES training.
\paragraph{Synthetic (Conventional)} Synthetic learner essays generated by generators fine-tuned directly on learner essays without error annotations, with the same prompt and label conditioning as the proposed approach. This condition represents a conventional synthetic essay generation setting without explicit supervision over learner error patterns, similar to prior proficiency-conditioned generation approaches \cite{chen-etal-2026-cpt, do2026swimstudentwritingsimulation}.

The data conditions are matched by prompt and target label, i.e., each synthetic instance is generated using the same prompt and proficiency level or score as its corresponding authentic training instance. For each condition, the AES scorer is fine-tuned under four training set sizes: 50, 100, and 200 samples, and the full training split size. For ELLIPSE, the full training split setting is instead replaced with a size-matched subset of 2,000 samples rather than the full size of 4,536 samples, as generating a synthetic dataset at this scale would require approximately 40 hours on an NVIDIA L4 GPU, given its relatively long average essay length of 433 words. We consider this sufficient to assess whether the advantage of error-tag supervision persists beyond the low-resource settings while remaining computationally feasible. Accordingly, all data conditions in this setting use 2,000 samples to ensure fair comparisons. 

We use BERT \cite{devlin-etal-2019-bert} as the sole AES scorer across experiments, as it offers a well-established transformer-based baseline in AES research \cite{schmalz-brutti-2021-automatic, wang-etal-2022-use, elks-2021-using-transfer}. Evaluating multiple architectures is beyond the scope of this study, as our primary objective is to compare the quality and downstream utility of the different training data conditions rather than the relative performance of alternative scoring models. For each corpus, BERT is fine-tuned separately under each of the above data conditions and training-set sizes. The resulting AES models are then evaluated on the same held-out authentic test split for that corpus. This ensures that all data conditions are compared under an identical evaluation setting. We use multiple sampling seeds to improve the robustness of comparisons in low-resource settings ($n={50,100,200}$), in which five subsets of size $n$ are independently constructed from the training split using different random seeds and proportional stratified sampling. Performance is reported as the mean and standard deviation over the five runs. All AES experiments are conducted under a prompt-unaware setting, with only the essay text provided as input. Further details of the AES experiments are provided in Appendix~\ref{sec:AES_details}.

\subsection{Results and Discussion}
\subsubsection{CLC FCE}

We treat AES on the CLC FCE corpus as a regression task, using the provided holistic essay score as the target label. This score ranges from 0 to 20. Accordingly, we evaluate model performance using Mean Absolute Error (MAE), Root Mean Squared Error (RMSE), Spearman's rank correlation, and Quadratic Weighted Kappa (QWK). Table \ref{tab:fce_results} presents the results of this experiment across the different data conditions. 

\begin{table*}[t]
\centering
\small
\caption{AES performance across data conditions on CLC FCE. CLC FCE's scores ranges from 0 to 20.  Panel A reports the full-data setting; Panel B reports mean ± standard deviation over five seeds for \(n=\{50,100,200\}\). Best performance within each setting is shown in bold. CLC FCE's scores range from 0 to 20.}
\label{tab:fce_results}
\renewcommand{\arraystretch}{1.12}
\begin{tabular}{clcccc}

\hline

\multicolumn{6}{l}{Panel A: Full-data setting ($n=1725$)} \\[2pt]
\hline

& \textbf{Data}
& \textbf{MAE $\downarrow$}
& \textbf{RMSE $\downarrow$}
& \textbf{Spearman $\uparrow$}
& \textbf{QWK $\uparrow$} \\

\hline

& Authentic & 1.8417 & 2.4045 & \textbf{0.6768} & \textbf{0.5850} \\

& Synthetic (Conventional) & 1.9749 & 2.5066 & 0.6079 & 0.5087 \\

& Synthetic (Proposed) & \textbf{1.8062} & \textbf{2.3510} & 0.6004 & 0.5105 \\

\hline

\multicolumn{6}{l}{Panel B: Low-resource settings} \\[2pt]
\hline

\textbf{$n$} & \textbf{Data} & \textbf{MAE $\downarrow$} & \textbf{RMSE $\downarrow$} & \textbf{Spearman $\uparrow$} & \textbf{QWK $\uparrow$} \\

\hline

& Authentic & $2.2680 \pm 0.0427$ & $2.8661 \pm 0.0630$ & $\mathbf{0.4288 \pm 0.0461}$ & $\mathbf{0.3088 \pm 0.0384}$ \\

50 & Synthetic (Conventional) & $\mathbf{2.2380 \pm 0.1399}$ & $\mathbf{2.8356 \pm 0.1654}$ & $0.2876 \pm 0.1097$ & $0.1905 \pm 0.0738$ \\

& Synthetic (Proposed) & $2.2524 \pm 0.1681$ & $2.8666 \pm 0.2011$ & $0.2244 \pm 0.2188$ & $0.1583 \pm 0.1439$ \\

\hline

& Authentic & $2.1431 \pm 0.0924$ & $2.7337 \pm 0.1081$ & $\mathbf{0.5155 \pm 0.0448}$ & $\mathbf{0.3858 \pm 0.0365}$ \\

100 & Synthetic (Conventional) & $\mathbf{2.1024 \pm 0.0834}$ & $\mathbf{2.7034 \pm 0.0858}$ & $0.4345 \pm 0.0632$ & $0.3537 \pm 0.0489$ \\

& Synthetic (Proposed) & $2.1816 \pm 0.1938$ & $2.7864 \pm 0.2336$ & $0.3416 \pm 0.1455$ & $0.2724 \pm 0.1318$ \\

\hline

& Authentic & $2.0895 \pm 0.1193$ & $2.6719 \pm 0.1229$ & $\mathbf{0.5721 \pm 0.0177}$ & $\mathbf{0.4564 \pm 0.0359}$ \\

200 & Synthetic (Conventional) & $2.1466 \pm 0.0888$ & $2.7426 \pm 0.0946$ & $0.4938 \pm 0.0333$ & $0.4073 \pm 0.0241$ \\

& Synthetic (Proposed) & $\mathbf{2.0314 \pm 0.1379}$ & $\mathbf{2.6231 \pm 0.1315}$ & $0.4890 \pm 0.0930$ & $0.4200 \pm 0.1031$ \\

\hline
\end{tabular}

\end{table*}

The results demonstrate a clear advantage for the proposed error-tagged synthetic data once sufficient training data are available. In the full-data setting, the proposed approach achieves the best MAE and RMSE, outperforming not only the conventional synthetic baseline but also the authentic-data baseline by a small margin. On Spearman correlation and QWK, the proposed approach performs comparably to the conventional synthetic baseline, with a slightly lower Spearman but higher QWK. Interestingly, the advantage of the proposed approach is more pronounced for MAE and RMSE than for Spearman correlation and QWK. This suggests that, on CLC FCE, error-tag supervision primarily improves the accuracy of the predicted scores rather than their relative ranking or ordinal agreement.

Under extremely low-resource settings, however, this advantage is not observed. With 50 and 100 training samples, neither the proposed synthetic data nor the authentic data outperform the conventional synthetic baseline on MAE and RMSE. Since the authentic data also fail to provide an advantage on these metrics, this suggests that the very small training-set size may constrain AES performance regardless of data source. The advantage emerges at $n=200$, where the proposed synthetic data outperform the conventional baseline on both MAE and RMSE, perform similarly on Spearman correlation, and achieve a slightly higher QWK. Overall, these results indicate that error-tag supervision provides a tangible benefit for downstream AES on CLC FCE, with the advantage over conventional synthetic data becoming more apparent from 200 training samples onward.

\subsubsection{Write \& Improve (W\&I)}
We treat AES on the W\&I corpus as a classification task, specifically CEFR classification of the writer's proficiency level. We therefore use Accuracy, Macro-F1, Weighted-F1, and QWK to assess the performance on this dataset. Table \ref{tab:wi_results} presents the results.

\begin{table*}[t]
\centering
\small
\caption{AES performance across data conditions on Write \& Improve. Panel A reports the full-data setting; Panel B reports mean ± standard deviation over five seeds for \(n=\{50,100,200\}\). Best performance within each setting is shown in bold.}
\label{tab:wi_results}
\renewcommand{\arraystretch}{1.12}
\begin{tabular}{clcccc}

\hline

\multicolumn{6}{l}{Panel A: Full-data setting ($n=3480$)} \\[2pt]
\hline

& \textbf{Data}
& \textbf{Accuracy $\uparrow$}
& \textbf{Macro-F1 $\uparrow$}
& \textbf{Weighted-F1 $\uparrow$}
& \textbf{QWK $\uparrow$} \\

\hline

& Authentic & \textbf{0.7273} & \textbf{0.4756} & \textbf{0.7281} & \textbf{0.8325} \\

& Synthetic (Conventional) & 0.5876 & 0.3842 & 0.5971 & 0.7504 \\

& Synthetic (Proposed) & 0.6983 & 0.4576 & 0.7004 & 0.8209 \\

\hline

\multicolumn{6}{l}{Panel B: Low-resource settings} \\[2pt]
\hline

\textbf{$n$} & \textbf{Data} & \textbf{Accuracy $\uparrow$} & \textbf{Macro-F1 $\uparrow$} & \textbf{Weighted-F1 $\uparrow$} & \textbf{QWK $\uparrow$} \\

\hline

& Authentic & $\mathbf{0.4614 \pm 0.0363}$ & $\mathbf{0.2068 \pm 0.0309}$ & $\mathbf{0.3899 \pm 0.0527}$ & $\mathbf{0.3949 \pm 0.1147}$ \\

50 & Synthetic (Conventional) & $0.4498 \pm 0.0494$ & $0.1927 \pm 0.0366$ & $0.3711 \pm 0.0652$ & $0.3484 \pm 0.1135$ \\

& Synthetic (Proposed) & $0.4406 \pm 0.0295$ & $0.1856 \pm 0.0165$ & $0.3530 \pm 0.0297$ & $0.3658 \pm 0.0746$ \\

\hline

& Authentic & $\mathbf{0.5647 \pm 0.0296}$ & $\mathbf{0.2910 \pm 0.0207}$ & $\mathbf{0.5379 \pm 0.0373}$ & $\mathbf{0.6342 \pm 0.0422}$ \\

100 & Synthetic (Conventional) & $0.5209 \pm 0.0379$ & $0.2552 \pm 0.0299$ & $0.4821 \pm 0.0555$ & $0.5190 \pm 0.0910$ \\

& Synthetic (Proposed) & $0.5101 \pm 0.0287$ & $0.2457 \pm 0.0147$ & $0.4598 \pm 0.0315$ & $0.5307 \pm 0.0160$ \\

\hline

& Authentic & $\mathbf{0.6451 \pm 0.0202}$ & $\mathbf{0.4044 \pm 0.0277}$ & $\mathbf{0.6408 \pm 0.0242}$ & $\mathbf{0.7549 \pm 0.0259}$ \\

200 & Synthetic (Conventional) & $0.5868 \pm 0.0185$ & $0.3664 \pm 0.0111$ & $0.5819 \pm 0.0180$ & $0.6867 \pm 0.0167$ \\

& Synthetic (Proposed) & $0.5863 \pm 0.0400$ & $0.3550 \pm 0.0430$ & $0.5670 \pm 0.0517$ & $0.6630 \pm 0.0543$ \\

\hline
\end{tabular}

\end{table*}

Under the full-data setting, the proposed synthetic data consistently outperform the conventional synthetic baseline across all evaluation metrics. In particular, the proposed approach improves Accuracy from 0.5876 to 0.6983, Macro-F1 from 0.3842 to 0.4576, Weighted-F1 from 0.5971 to 0.7004, and QWK from 0.7504 to 0.8209. Its performance is also largely comparable to that of the authentic-data baseline, with only small differences across all four metrics. Notably, the proposed synthetic condition approaches the authentic baseline particularly closely on QWK, suggesting that the generated data preserve useful information about the ordinal relationships between CEFR proficiency levels. The magnitude of this improvement is notably larger than that observed on CLC FCE and ELLIPSE, which may reflect differences in the task formulation or other dataset-specific characteristics.

Under the low-resource settings, however, the advantage of the proposed approach over the conventional synthetic baseline does not persist consistently. At 50 and 100 samples, the two synthetic conditions achieve broadly similar performance, with the proposed approach obtaining slightly higher QWK but slightly lower scores on the remaining metrics. At $n=200$, the proposed and conventional approaches achieve nearly identical mean accuracy, while the conventional baseline performs slightly better on Macro-F1, Weighted-F1, and QWK. Overall, these results suggest that the benefit of error-tag supervision on W\&I becomes apparent at the full-data scale, but provides no consistent improvement over conventional synthetic data in the low-resource settings examined.

\subsubsection{ELLIPSE}
Similar to the CLC FCE corpus, we treat AES on ELLIPSE as a regression task, using the essay's holistic score as the target variable. The scores range from 1.0 to 5.0. Accordingly, we evaluate model performance using MAE, RMSE, Spearman's correlation, and QWK. Table \ref{tab:ellipse_results} presents the results across the three data conditions.

\begin{table*}[t]
\centering
\small
\caption{AES performance across data conditions on ELLIPSE. Panel A reports the larger-data setting with 2000 samples; Panel B reports mean ± standard deviation over five seeds for \(n=\{50,100,200\}\) samples. Best performance within each setting is shown in bold. ELLIPSE's scores range from 1.0 to 5.0}
\label{tab:ellipse_results}
\renewcommand{\arraystretch}{1.12}
\begin{tabular}{clcccc}

\hline

\multicolumn{6}{l}{Panel A: Larger-data setting ($n=2000$)} \\[2pt]
\hline

& \textbf{Data}
& \textbf{MAE $\downarrow$}
& \textbf{RMSE $\downarrow$}
& \textbf{Spearman $\uparrow$}
& \textbf{QWK $\uparrow$} \\

\hline

& Authentic & \textbf{0.3614} & \textbf{0.4489} & \textbf{0.7282} & \textbf{0.7010} \\

& Synthetic (Conventional) & 0.4389 & 0.5420 & 0.6860 & 0.5963 \\

& Synthetic (Proposed) & 0.4330 & 0.5321 & 0.6900 & 0.6116 \\

\hline

\multicolumn{6}{l}{Panel B: Low-resource settings} \\[2pt]
\hline

\textbf{$n$} & \textbf{Data} & \textbf{MAE $\downarrow$} & \textbf{RMSE $\downarrow$} & \textbf{Spearman $\uparrow$} & \textbf{QWK $\uparrow$} \\

\hline

& Authentic & $0.4636 \pm 0.0418$ & $0.5725 \pm 0.0472$ & $\mathbf{0.6484 \pm 0.0114}$ & $\mathbf{0.5187 \pm 0.0442}$ \\

50 & Synthetic (Conventional) & $\mathbf{0.4323 \pm 0.0113}$ & $0.5427 \pm 0.0143$ & $0.5989 \pm 0.0204$ & $0.5112 \pm 0.0316$ \\

& Synthetic (Proposed) & $0.4323 \pm 0.0259$ & $\mathbf{0.5411 \pm 0.0299}$ & $0.5982 \pm 0.0507$ & $0.5123 \pm 0.0600$ \\

\hline

& Authentic & $\mathbf{0.4109 \pm 0.0245}$ & $\mathbf{0.5104 \pm 0.0309}$ & $\mathbf{0.6609 \pm 0.0154}$ & $\mathbf{0.5860 \pm 0.0341}$ \\

100 & Synthetic (Conventional) & $0.4484 \pm 0.0320$ & $0.5569 \pm 0.0359$ & $0.6152 \pm 0.0123$ & $0.5289 \pm 0.0176$ \\

& Synthetic (Proposed) & $0.4507 \pm 0.0363$ & $0.5585 \pm 0.0434$ & $0.6279 \pm 0.0252$ & $0.5373 \pm 0.0478$ \\

\hline

& Authentic & $\mathbf{0.3908 \pm 0.0182}$ & $\mathbf{0.4874 \pm 0.0224}$ & $\mathbf{0.6713 \pm 0.0154}$ & $\mathbf{0.6265 \pm 0.0286}$ \\

200 & Synthetic (Conventional) & $0.4940 \pm 0.0386$ & $0.6032 \pm 0.0433$ & $0.6455 \pm 0.0190$ & $0.5504 \pm 0.0281$ \\

& Synthetic (Proposed) & $0.4709 \pm 0.0316$ & $0.5816 \pm 0.0379$ & $0.6505 \pm 0.0165$ & $0.5618 \pm 0.0308$ \\

\hline
\end{tabular}

\end{table*}

As discussed earlier, ELLIPSE uses a larger-data setting of 2,000 training samples rather than the full 4,536 samples available in the training split, due to the substantial computational cost of generating synthetic datasets at the full scale. Under this setting, the model trained on the proposed synthetic data outperforms the conventional synthetic baseline across all evaluation metrics, although the magnitude of the improvement is modest. Nevertheless, both synthetic conditions still perform noticeably below the authentic-data baseline, indicating that while error-tag supervision improves the downstream utility of synthetic ELLIPSE essays, a substantial gap remains between synthetic and authentic training data. 

Under the low-resource settings, the instability observed in earlier experiments on W\&I and CLC FCE also persists on ELLIPSE. At $n=50$, the results are highly mixed: the conventional synthetic baseline achieves the lowest MAE, although the difference from the proposed condition is negligible and both values round to 0.4323, while the proposed approach achieves the lowest RMSE. In contrast, the authentic data baseline performs best on both Spearman correlation and QWK. Notably, the difference in MAE between the two synthetic conditions is extremely small, further illustrating the instability of the comparison at this training-set size. Taken together with the similarly mixed low-resource results observed on the other corpora, this suggests that performance at $n=50$ may be strongly affected by the extremely limited amount of training data. A clearer ordering emerges at $n=100$, where the authentic-data baseline outperforms both synthetic conditions across all four metrics. However, in this setting, neither synthetic approach consistently dominates the other: the conventional baseline performs better on MAE and RMSE, whereas the proposed approach achieves a higher Spearman correlation and QWK. A clearer advantage for the proposed approach emerges at $n=200$, where it outperforms the conventional synthetic baseline across all four evaluation metrics, although the authentic-data baseline continues to achieve the best overall performance.

The ELLIPSE results are noteworthy because the error annotations used in this corpus are automatically derived using GECToR-2024 and ERRANT rather than provided through manual corpus annotation. Although the observed advantage of the proposed approach over the conventional synthetic baseline is modest, it nevertheless suggests that the methodology may not be limited to learner corpora that already contain explicit error annotations or corrected reference texts. Instead, comparable supervision may potentially be constructed for datasets that provide only original learner essays using existing GEC/GED tools, thereby extending the applicability of the approach to a broader range of AES datasets. Although this finding is based on a single automatically annotated corpus, it provides preliminary evidence that manually error-annotated resources are not a strict prerequisite for applying the proposed methodology.

\section{Conclusion}
This study introduces a simple approach to generating synthetic learner essays for Automated Essay Scoring (AES) by fine-tuning an LLM to produce error-annotated texts similar to those commonly found in corpora developed for Grammatical Error Correction (GEC) and Grammatical Error Detection (GED). We evaluate essays generated by our approach against two baselines: authentic essays and conventionally generated synthetic essays. Under larger-data settings, AES scorers trained on essays generated using the proposed approach generally outperform those trained on conventionally generated synthetic essays across datasets and evaluation settings. Under the full-data setting on CLC FCE, the proposed synthetic data even yield performance comparable to or slightly better than authentic learner data. However, results remain mixed in the extremely low-resource settings with only 50 or 100 training samples. Qualitative and quantitative analyses further show that the generated essays exhibit error profiles broadly resembling those of authentic learner essays at the corresponding proficiency level or score, although weaknesses such as repetition, semantic inconsistencies, and mismatches in error density remain.

These findings suggest that improving the error characteristics of synthetic learner essays for AES does not necessarily require a sophisticated post-hoc error-injection pipeline. A simple change to the generator's supervision signal can improve the usefulness of the resulting synthetic data. Importantly, this supervision can be obtained from existing error-annotated learner corpora developed for GEC and GED, extending the value of these resources beyond their original tasks. The ELLIPSE experiments further suggest that the approach is not limited to corpora with pre-existing manual error annotations, as automatically derived error tags can also provide useful supervision, making the method applicable to a broader range of learner-writing datasets. Potentially, these findings also motivate future corpus development to include error annotations where feasible, given their observed utility in our experiments as a supervision signal.

\section*{Limitations}\label{sec:limitation}
Several limitations should be addressed in the present study. First, although the proposed approach generally improves downstream AES performance, its advantage is less consistent under the extremely low-resource settings with only 50 or 100 training samples. The results therefore do not suggest that error-tag supervision provides a universal benefit across all data settings.

Second, the error annotations used for ELLIPSE are automatically derived using GEC and GED tools rather than manually annotated by human experts. As a result, these annotations may contain incorrect corrections, missed errors, or inaccurate error types, introducing noise into the supervision signal. While the results show that such automatically derived annotations can still be useful, their quality may influence both the generated error patterns and downstream AES performance.

Third, the generated essays do not perfectly reproduce authentic learner writing. The qualitative and quantitative analyses reveal remaining discrepancies in error density and proficiency-specific error patterns, as well as generation weaknesses such as repetition, semantic inconsistencies, and over-correction. The proposed approach should therefore be viewed as improving, rather than fully reproducing, authentic learner-error characteristics.

Finally, the experiments are limited to three English learner corpora and the model architectures considered in this study. The proposed approach also requires substantially more time for fine-tuning and data generation than conventional generation, potentially due in part to the longer sequences introduced by the additional error-markup tokens (see Appendix \ref{sec:gen_details}). This overhead becomes particularly pronounced for corpora containing longer essays. Consequently, the larger-data experiment on ELLIPSE uses a size-matched subset of 2,000 training samples rather than the complete training split. Further evidence is therefore needed before the findings can be generalized to other learner populations, languages, generator models, AES architectures, or substantially larger-scale generation settings.
\section*{Ethical Considerations}
The proposed approach is intended for research on synthetic learner data and Automated Essay Scoring. However, because it is designed to generate texts that reproduce learner-like error patterns and proficiency characteristics, it could potentially be misused to produce LLM-generated essays that appear to have been written by genuine language learners, with the aim of misleading human graders or automated assessment systems. Such use could facilitate academic misconduct in educational settings. This risk should therefore be considered when deploying or releasing systems based on the proposed method.
Regarding the data used in this study, all three corpora are restricted to non-commercial use, with CLC FCE and W\&I further limited to non-commercial research and educational purposes. Any future work reproducing or extending this study using these corpora should therefore comply with their respective licensing and redistribution requirements. Written permission was obtained from Cambridge University Press \& Assessment to include the generated CLC FCE and W\&I essays in the appendix. ELLIPSE is distributed under the CC BY-NC-SA 4.0 licence, under which the corresponding materials are used and presented in accordance with its attribution, non-commercial use, and ShareAlike requirements.
\section*{Acknowledgement}
We thank Cambridge University Press \& Assessment for providing access to the CLC FCE and Write \& Improve datasets, and for approving our requests to include generated samples in the appendix of this paper.
\bibliography{custom}

@inproceedings{yoo-etal-2025-dress,
    title = "{DRE}s{S}: Dataset for Rubric-based Essay Scoring on {EFL} Writing",
    author = "Yoo, Haneul  and
      Han, Jieun  and
      Ahn, So-Yeon  and
      Oh, Alice",
    editor = "Che, Wanxiang  and
      Nabende, Joyce  and
      Shutova, Ekaterina  and
      Pilehvar, Mohammad Taher",
    booktitle = "Proceedings of the 63rd Annual Meeting of the Association for Computational Linguistics (Volume 1: Long Papers)",
    month = jul,
    year = "2025",
    address = "Vienna, Austria",
    publisher = "Association for Computational Linguistics",
    url = "https://aclanthology.org/2025.acl-long.659/",
    doi = "10.18653/v1/2025.acl-long.659",
    pages = "13439--13454",
    ISBN = "979-8-89176-251-0"
}

@inproceedings{qwaider-etal-2025-enhancing,
    title = "Enhancing {A}rabic Automated Essay Scoring with Synthetic Data and Error Injection",
    author = "Qwaider, Chatrine  and
      Alhafni, Bashar  and
      Chirkunov, Kirill  and
      Habash, Nizar  and
      Briscoe, Ted",
    editor = {Kochmar, Ekaterina  and
      Alhafni, Bashar  and
      Bexte, Marie  and
      Burstein, Jill  and
      Horbach, Andrea  and
      Laarmann-Quante, Ronja  and
      Tack, Ana{\"i}s  and
      Yaneva, Victoria  and
      Yuan, Zheng},
    booktitle = "Proceedings of the 20th Workshop on Innovative Use of NLP for Building Educational Applications (BEA 2025)",
    month = jul,
    year = "2025",
    address = "Vienna, Austria",
    publisher = "Association for Computational Linguistics",
    url = "https://aclanthology.org/2025.bea-1.40/",
    doi = "10.18653/v1/2025.bea-1.40",
    pages = "549--563",
    ISBN = "979-8-89176-270-1"
}

@Article{Park-2022-essayGAN,
AUTHOR = {Park, Yo-Han and Choi, Yong-Seok and Park, Cheon-Young and Lee, Kong-Joo},
TITLE = {EssayGAN: Essay Data Augmentation Based on Generative Adversarial Networks for Automated Essay Scoring},
JOURNAL = {Applied Sciences},
VOLUME = {12},
YEAR = {2022},
NUMBER = {12},
ARTICLE-NUMBER = {5803},
URL = {https://www.mdpi.com/2076-3417/12/12/5803},
ISSN = {2076-3417},
DOI = {10.3390/app12125803}
}

@inproceedings{Zhang-et-al-2026,
author="Zhang, Yuli
and Luo, Yihan
and Hua, Shenglu
and Yu, Yuanlong
and Cui, Bokai",
editor="Taniguchi, Tadahiro
and Leung, Chi Sing Andrew
and Kozuno, Tadashi
and Yoshimoto, Junichiro
and Mahmud, Mufti
and Doborjeh, Maryam
and Doya, Kenji",
title="Data Augmentation for Automated Essay Scoring Using Large Language Models and In-Context Learning",
booktitle="Neural Information Processing",
year="2026",
publisher="Springer Nature Singapore",
address="Singapore",
pages="274--288",
isbn="978-981-95-4091-4",
url="https://link.springer.com/chapter/10.1007/978-981-95-4091-4_19"
}

@InProceedings{nam-2027-prompt,
author="Nam, Sungjin",
editor="Blanchard, Emmanuel G.
and Chen, Guanliang
and Chi, Min
and Isotani, Seiji",
title="Prompt Optimization with Verifiable Rewards for Synthetic Essay Generation",
booktitle="Artificial Intelligence in Education",
year="2027",
publisher="Springer Nature Switzerland",
address="Cham",
pages="125--139",
isbn="978-3-032-29744-0",
doi="10.1007/978-3-032-29744-0_9"
}

@inproceedings{chen-etal-2026-cpt,
    title = "{CPT}-Agent: A Cognitive Process Theory-driven Framework for Student Simulation in Writing Development",
    author = "Chen, Yuhan  and
      Shen, Zizhuo  and
      Cheng, Miaomiao  and
      Han, Xu  and
      Gong, Jiefu  and
      Wang, Shijin  and
      Song, Wei",
    editor = "Liakata, Maria  and
      Moreira, Viviane P.  and
      Zhang, Jiajun  and
      Jurgens, David",
    booktitle = "Proceedings of the 64th Annual Meeting of the {A}ssociation for {C}omputational {L}inguistics (Volume 1: Long Papers)",
    month = jul,
    year = "2026",
    address = "San Diego, California, United States",
    publisher = "Association for Computational Linguistics",
    url = "https://aclanthology.org/2026.acl-long.846/",
    doi = "10.18653/v1/2026.acl-long.846",
    pages = "18596--18616",
    ISBN = "979-8-89176-390-6"
}

@inproceedings{yannakoudakis-etal-2011-new,
    title = "A New Dataset and Method for Automatically Grading {ESOL} Texts",
    author = "Yannakoudakis, Helen  and
      Briscoe, Ted  and
      Medlock, Ben",
    editor = "Lin, Dekang  and
      Matsumoto, Yuji  and
      Mihalcea, Rada",
    booktitle = "Proceedings of the 49th Annual Meeting of the Association for Computational Linguistics: Human Language Technologies",
    month = jun,
    year = "2011",
    address = "Portland, Oregon, USA",
    publisher = "Association for Computational Linguistics",
    url = "https://aclanthology.org/P11-1019/",
    pages = "180--189"
}

@misc{nicholls-etal-2024-write-improve,
    author       = {Nicholls, Diane and Caines, Andrew and Buttery, Paula},
    title        = {The {Write \& Improve Corpus 2024}: Error-Annotated and {CEFR}-Labelled Essays by Learners of English},
    year         = {2024},
    howpublished = {Apollo -- University of Cambridge Repository},
    doi          = {10.17863/CAM.112997},
    url          = {https://doi.org/10.17863/CAM.112997}
}

@inproceedings{bryant-etal-2017-automatic,
    title = {Automatic Annotation and Evaluation of Error Types for Grammatical Error Correction},
    author = {Bryant, Christopher and Felice, Mariano and Briscoe, Ted},
    booktitle = {Proceedings of the 55th Annual Meeting of the Association for Computational Linguistics (Volume 1: Long Papers)},
    year = {2017},
    address = {Vancouver, Canada},
    publisher = {Association for Computational Linguistics},
    pages = {793--805},
    doi = {10.18653/v1/P17-1074},
    url = {https://aclanthology.org/P17-1074/}
}

@article{Zhang_Badola_Johnson_Li_2026,
title={Augmenting {AI} scoring of essays with {GPT-generated responses}}, volume={17}, url={https://www.jowr.org/jowr/article/view/1777}, DOI={10.17239/jowr-2026.17.03.06}, 
abstractNote={
In this study, we examine the feasibility of augmenting student-written essays with those generated by large language models (LLMs) for scoring essays. We found that with correct instructions, generative AI systems such as GPT-4 and GPT-4o can generate essays similar to those written by students in terms of surface-level linguistic features, although material differences may still exist. Systematic analyses revealed that scoring models trained with synthetic data perform comparably to models trained using student essays, but the performance varies across prompts and the sizes of the model training sample. The augmented models could alleviate large discrepancies between human and AI scores on the subgroup level that may be introduced by a lack of training samples for a particular subgroup or due to inherent biases in LLMs. We also explored an established method – DecompX – on token importance to identify and explain AI predictions. Future research directions and limitations of this study are also discussed.
},
number={3}, 
journal={Journal of Writing Research}, 
author={Zhang, Mo and Badola, Akshay and Johnson, Matthew and Li, Chen}, year={2026},
month={Feb.} 
}

@inproceedings{QLoRA,
 author = {Dettmers, Tim and Pagnoni, Artidoro and Holtzman, Ari and Zettlemoyer, Luke},
 booktitle = {Advances in Neural Information Processing Systems},
 doi = {10.52202/075280-0441},
 editor = {A. Oh and T. Naumann and A. Globerson and K. Saenko and M. Hardt and S. Levine},
 pages = {10088--10115},
 publisher = {Curran Associates, Inc.},
 title = {QLoRA: Efficient Finetuning of Quantized LLMs},
 url = {https://proceedings.neurips.cc/paper_files/paper/2023/file/1feb87871436031bdc0f2beaa62a049b-Paper-Conference.pdf},
 volume = {36},
 year = {2023}
}

@inproceedings{LoRA,
author = {Hu, Edward J. and Shen, Yelong and Wallis, Phillip and Allen-Zhu, Zeyuan and Yuanzhi  Li and Wang, Shean and Wang, Lu and Chen, Weizhu},
title = {LoRA: Low-Rank Adaptation of Large Language Models},
booktitle = {ICLR 2022},
year = {2022},
month = {April},
url = {https://www.microsoft.com/en-us/research/publication/lora-low-rank-adaptation-of-large-language-models/},
}

@inproceedings{elks-2021-using-transfer,
    title = "Using Transfer Learning to Automatically Mark {L}2 Writing Texts",
    author = "Elks, Tim",
    editor = "Djabri, Souhila  and
      Gimadi, Dinara  and
      Mihaylova, Tsvetomila  and
      Nikolova-Koleva, Ivelina",
    booktitle = "Proceedings of the Student Research Workshop Associated with RANLP 2021",
    month = sep,
    year = "2021",
    address = "Online",
    publisher = "INCOMA Ltd.",
    url = "https://aclanthology.org/2021.ranlp-srw.8/",
    pages = "51--57"
}

@inproceedings{wang-etal-2022-use,
    title = "On the Use of {Bert} for {Automated Essay Scoring}: Joint Learning of Multi-Scale Essay Representation",
    author = "Wang, Yongjie  and
      Wang, Chuang  and
      Li, Ruobing  and
      Lin, Hui",
    editor = "Carpuat, Marine  and
      de Marneffe, Marie-Catherine  and
      Meza Ruiz, Ivan Vladimir",
    booktitle = "Proceedings of the 2022 Conference of the North American Chapter of the Association for Computational Linguistics: Human Language Technologies",
    month = jul,
    year = "2022",
    address = "Seattle, United States",
    publisher = "Association for Computational Linguistics",
    url = "https://aclanthology.org/2022.naacl-main.249/",
    doi = "10.18653/v1/2022.naacl-main.249",
    pages = "3416--3425"
}

@inproceedings{schmalz-brutti-2021-automatic,
    title = "Automatic Assessment of {E}nglish {CEFR} Levels Using {BERT} Embeddings",
    author = "Schmalz, Veronica Juliana  and
      Brutti, Alessio",
    editor = "Fersini, Elisabetta  and
      Passarotti, Marco  and
      Patti, Viviana",
    booktitle = "Proceedings of the Eighth Italian Conference on Computational Linguistics (CLiC-it 2021)",
    month = jun,
    year = "2021",
    address = "Milan, Italy",
    publisher = "CEUR Workshop Proceedings",
    url = "https://aclanthology.org/2021.clicit-1.45/",
    pages = "295--301",
    ISBN = "979-12-80136-94-7"
}

@inproceedings{devlin-etal-2019-bert,
    title = "{BERT}: Pre-training of Deep Bidirectional Transformers for Language Understanding",
    author = "Devlin, Jacob  and
      Chang, Ming-Wei  and
      Lee, Kenton  and
      Toutanova, Kristina",
    editor = "Burstein, Jill  and
      Doran, Christy  and
      Solorio, Thamar",
    booktitle = "Proceedings of the 2019 Conference of the North {A}merican Chapter of the Association for Computational Linguistics: Human Language Technologies, Volume 1 (Long and Short Papers)",
    month = jun,
    year = "2019",
    address = "Minneapolis, Minnesota",
    publisher = "Association for Computational Linguistics",
    url = "https://aclanthology.org/N19-1423/",
    doi = "10.18653/v1/N19-1423",
    pages = "4171--4186"
}

@misc{yang2024qwen2technicalreport,
      title={Qwen2 Technical Report}, 
      author={An Yang and Baosong Yang and Binyuan Hui and Bo Zheng and Bowen Yu and Chang Zhou and Chengpeng Li and Chengyuan Li and Dayiheng Liu and Fei Huang and Guanting Dong and Haoran Wei and Huan Lin and Jialong Tang and Jialin Wang and Jian Yang and Jianhong Tu and Jianwei Zhang and Jianxin Ma and Jianxin Yang and Jin Xu and Jingren Zhou and Jinze Bai and Jinzheng He and Junyang Lin and Kai Dang and Keming Lu and Keqin Chen and Kexin Yang and Mei Li and Mingfeng Xue and Na Ni and Pei Zhang and Peng Wang and Ru Peng and Rui Men and Ruize Gao and Runji Lin and Shijie Wang and Shuai Bai and Sinan Tan and Tianhang Zhu and Tianhao Li and Tianyu Liu and Wenbin Ge and Xiaodong Deng and Xiaohuan Zhou and Xingzhang Ren and Xinyu Zhang and Xipin Wei and Xuancheng Ren and Xuejing Liu and Yang Fan and Yang Yao and Yichang Zhang and Yu Wan and Yunfei Chu and Yuqiong Liu and Zeyu Cui and Zhenru Zhang and Zhifang Guo and Zhihao Fan},
      year={2024},
      eprint={2407.10671},
      archivePrefix={arXiv},
      primaryClass={cs.CL},
      url={https://arxiv.org/abs/2407.10671}, 
}

@article{ETAAT2026100258,
title = {Exploring linguistic fingerprints in human and {AI}-generated texts: An {NLP}-based approach in second language writing},
journal = {Ampersand},
volume = {16},
pages = {100258},
year = {2026},
issn = {2215-0390},
doi = {https://doi.org/10.1016/j.amper.2026.100258},
url = {https://www.sciencedirect.com/science/article/pii/S2215039026000056},
author = {Fatemeh Etaat}
}

@article{ellipse,
  author    = {Crossley, Scott and Tian, Yu and Baffour, Perpetual and Franklin, Alex and Kim, Youngmeen and Morris, Wesley and Benner, Meg and Picou, Aigner and Boser, Ulrich},
  title     = {The English Language Learner Insight, Proficiency and Skills Evaluation ({ELLIPSE}) Corpus},
  journal   = {International Journal of Learner Corpus Research},
  year      = {2023},
  volume    = {9},
  number    = {2},
  pages     = {248--269},
  doi       = {10.1075/ijlcr.22026.cro},
  url       = {https://doi.org/10.1075/ijlcr.22026.cro},
  publisher = {John Benjamins},
  issn      = {2215-1478}
}

@inproceedings{omelianchuk-etal-2024-pillars,
    title = "Pillars of Grammatical Error Correction: Comprehensive Inspection Of Contemporary Approaches In The Era of Large Language Models",
    author = "Omelianchuk, Kostiantyn  and
      Liubonko, Andrii  and
      Skurzhanskyi, Oleksandr  and
      Chernodub, Artem  and
      Korniienko, Oleksandr  and
      Samokhin, Igor",
    editor = {Kochmar, Ekaterina  and
      Bexte, Marie  and
      Burstein, Jill  and
      Horbach, Andrea  and
      Laarmann-Quante, Ronja  and
      Tack, Ana{\"i}s  and
      Yaneva, Victoria  and
      Yuan, Zheng},
    booktitle = "Proceedings of the 19th Workshop on Innovative Use of NLP for Building Educational Applications (BEA 2024)",
    month = jun,
    year = "2024",
    address = "Mexico City, Mexico",
    publisher = "Association for Computational Linguistics",
    url = "https://aclanthology.org/2024.bea-1.3/",
    pages = "17--33"
}

@article{bryant-etal-2023-grammatical,
    title = "Grammatical Error Correction: A Survey of the State of the Art",
    author = "Bryant, Christopher  and
      Yuan, Zheng  and
      Qorib, Muhammad Reza  and
      Cao, Hannan  and
      Ng, Hwee Tou  and
      Briscoe, Ted",
    journal = "Computational Linguistics",
    volume = "49",
    number = "3",
    month = sep,
    year = "2023",
    address = "Cambridge, MA",
    publisher = "MIT Press",
    url = "https://aclanthology.org/2023.cl-3.4/",
    doi = "10.1162/coli_a_00478",
    pages = "643--701"
}

@ARTICLE{JSD,
  author={Lin, J.},
  journal={IEEE Transactions on Information Theory}, 
  title={Divergence measures based on the Shannon entropy}, 
  year={1991},
  volume={37},
  number={1},
  pages={145-151},
  doi={10.1109/18.61115}}

@inproceedings{rei-etal-2017-artificial,
    title = "Artificial Error Generation with Machine Translation and Syntactic Patterns",
    author = "Rei, Marek  and
      Felice, Mariano  and
      Yuan, Zheng  and
      Briscoe, Ted",
    editor = "Tetreault, Joel  and
      Burstein, Jill  and
      Leacock, Claudia  and
      Yannakoudakis, Helen",
    booktitle = "Proceedings of the 12th Workshop on Innovative Use of {NLP} for Building Educational Applications",
    month = sep,
    year = "2017",
    address = "Copenhagen, Denmark",
    publisher = "Association for Computational Linguistics",
    url = "https://aclanthology.org/W17-5032/",
    doi = "10.18653/v1/W17-5032",
    pages = "287--292"
}

@misc{do2026swimstudentwritingsimulation,
      title={SWIM: Student Writing Simulation via Proficiency-Conditioned Generation}, 
      author={Heejin Do and Jakub Kontak and Mrinmaya Sachan},
      year={2026},
      eprint={2609.03215},
      archivePrefix={arXiv},
      primaryClass={cs.CL},
      url={https://arxiv.org/abs/2609.03215}, 
}

@article{kullback-leibler,
 ISSN = {00034851},
 URL = {http://www.jstor.org/stable/2236703},
 author = {S. Kullback and R. A. Leibler},
 journal = {The Annals of Mathematical Statistics},
 number = {1},
 pages = {79--86},
 publisher = {Institute of Mathematical Statistics},
 title = {On Information and Sufficiency},
 urldate = {2026-05-15},
 volume = {22},
 year = {1951}
}

\appendix

\section{Generator Training and Data Generation Details}\label{sec:gen_details}

In this appendix, we provide further details on the training process of the synthetic essay generators, including the system prompts and training hyperparameters. Table \ref{tab:system_prompt} presents the prompts used for generation. The three datasets use the same overall prompt structure, with minor dataset-specific adaptations to reflect their respective proficiency or scoring schemes. The conventional baseline described in Section \ref{sec:experimental_setup} uses the corresponding prompt for each dataset with the error-annotation instructions removed.

In addition, Table \ref{tab:generator_hyperparameters} reports the hyperparameters used to fine-tune the Qwen2-7B-Instruct generators. ELLIPSE differs from FCE and W\&I in training batch size, evaluation batch size, gradient accumulation, and maximum sequence length. Because ELLIPSE essays are substantially longer, the default batch settings resulted in out-of-memory errors during training. We therefore reduced both the training and evaluation batch sizes from 2 to 1 and increased gradient accumulation from 8 to 16, thereby maintaining the same effective batch size of 16. A larger maximum sequence length was also used for ELLIPSE to accommodate its longer essays.

Additionally, Table \ref{tab:computational_time} reports the approximate time required for generator fine-tuning and synthetic data generation. As discussed in Section \ref{sec:limitation}, the proposed approach requires considerably more training time than the conventional baseline, potentially due in part to the longer sequences introduced by the additional markup tokens. The increase is even more pronounced during data generation, with generation time approximately doubling for FCE and W\&I.

\begin{table*}[t]
\centering
\small
\begin{tabular}{p{0.96\textwidth}}
\hline
\textbf{System Prompt} \\
\hline

You generate authentic English learner writing for research.

\medskip
Follow the writing task and the requested target [TARGET].

\medskip
Write the entire response as a learner whose overall language ability is
representative of that [TARGET]. Match not only grammatical accuracy, but
also the learner's typical vocabulary, word choice, sentence structure,
fluency, organization, awkward phrasing, repetition, and level of
sophistication.

\medskip
Do not write a polished native-like essay and then artificially insert a few
errors. The whole response should naturally resemble learner writing at the
requested [TARGET].

\medskip
Preserve realistic learner-like grammatical, lexical, spelling, punctuation,
and usage errors where appropriate.

\medskip
Mark only genuine learner errors inline using:
\texttt{<ERR type="ERROR\_TYPE" cor="CORRECTION">erroneous text</ERR>}

\medskip
The \texttt{cor} attribute contains the intended corrected form of the
erroneous text.

\medskip
For errors involving a genuinely missing item, use:
\texttt{<ERR type="ERROR\_TYPE" cor="MISSING\_ITEM"><MISSING/></ERR>}

\medskip
For an unnecessary item that should simply be deleted, use an empty
correction:
\texttt{<ERR type="ERROR\_TYPE" cor="">unnecessary text</ERR>}

\medskip
Do not tag text merely because it is stylistically awkward if it is not
actually an error. Do not invent error labels unnecessarily.

\medskip
Return only the complete error-tagged learner essay.
\\

\hline
\end{tabular}

\caption{System prompt used for error-supervised synthetic essay generation.
[TARGET] denotes the dataset-specific proficiency signal: \emph{score} for
CLC FCE, \emph{CEFR level} for Write \& Improve, and \emph{holistic score}
for ELLIPSE.}
\label{tab:system_prompt}
\end{table*}

\begin{table}[t]
\centering
\small
\caption{Hyperparameters used for QLoRA fine-tuning of the synthetic essay generators.}
\label{tab:generator_hyperparameters}

\begin{tabular}{ll}
\hline
\multicolumn{2}{l}{\textbf{Panel A: Shared hyperparameters}} \\
\hline
\textbf{Hyperparameter} & \textbf{Value} \\
\hline
Base model & Qwen2-7B-Instruct \\
Hardware & A100 GPU \\
Number of epochs & 3 \\
Learning rate & $1\times10^{-4}$ \\
LR scheduler & Cosine \\
Warmup ratio & 0.05 \\
Weight decay & 0.01 \\
Effective batch size & 16 \\
Maximum gradient norm & 1.0 \\
LoRA rank ($r$) & 16 \\
LoRA $\alpha$ & 32 \\
LoRA dropout & 0.05 \\
LoRA target modules & All linear layers \\
Quantization & 4-bit NF4 \\
Double quantization & Yes \\
Compute precision & bfloat16 \\
Gradient checkpointing & Enabled \\
Sequence packing & Disabled \\
Loss & Completion-only causal LM \\
Evaluation frequency & Once per epoch \\
Model selection criterion & Dev loss \\
Random seed & 42 \\
\end{tabular}

\vspace{0.5em}

\begin{tabular}{lcc}
\hline
\multicolumn{3}{l}{\textbf{Panel B: Dataset-specific hyperparameters}} \\
\hline
\textbf{Hyperparameter} & \textbf{CLC FCE/W\&I} & \textbf{ELLIPSE} \\
\hline
Train batch size & 2 & 1 \\
Eval. batch size & 2 & 1 \\
Grad. accumulation & 8 & 16 \\
Max. sequence length & 2048 & 8192 \\
\hline
\end{tabular}

\end{table}

\begin{table}[t]
\centering
\small
\caption{Approximate wall-clock time for generator fine-tuning and synthetic essay generation.}
\label{tab:computational_time}
\begin{tabular}{llrr}
\toprule
Dataset & Method & Time & Increase \\
\midrule

\multicolumn{4}{l}{\textit{Training}} \\
\midrule
FCE
& Conventional & 32 min & -- \\
& Proposed     & 53 min & +66\% \\

W\&I
& Conventional & 50 min & -- \\
& Proposed     & 79 min & +59\% \\

ELLIPSE
& Conventional & 2 h 22 min & -- \\
& Proposed     & 3 h 36 min & +52\% \\

\midrule
\multicolumn{4}{l}{\textit{Generation}} \\
\midrule
FCE
& Conventional & 47 min 44 s & -- \\
& Proposed     & 1 h 45 min & +121.6\% \\

W\&I
& Conventional & 1 h 30 min & -- \\
& Proposed     & 2 h 58 min & +98.0\% \\

ELLIPSE
& Conventional & 12 h 38 min & -- \\
& Proposed     & 22 h 41 min & +79.5\% \\

\bottomrule
\end{tabular}
\end{table}

\section{Detailed Error Analyses}\label{sec:detailed_analyses}
\subsection{Example Generated Essays}
To complement the qualitative assessment in Section \ref{sec:analyses}, we provide one example essay from each generated dataset. The examples are drawn from the same 30-essay subsets used for manual inspection in the analysis presented in the main paper. We selected examples from intermediate proficiency levels or scores that exhibit typical generation behavior, including both learner-like errors and some of the imperfections observed during qualitative inspection. Table \ref{tab:example_essays} presents the examples for CLC FCE, W\&I, and ELLIPSE, respectively, from top to bottom. Written permission was obtained from the relevant dataset providers to include derived examples from CLC FCE and W\&I. ELLIPSE is publicly distributed under the CC BY-NC-SA 4.0 license and was used in accordance with its licensing terms.

\clearpage
\onecolumn

\begin{longtable}{@{}p{0.14\textwidth} p{0.82\textwidth}@{}}
\caption{Example essays generated using the proposed approach for CLC FCE, Write \& Improve, and ELLIPSE.}
\label{tab:example_essays} \\

\toprule
\multicolumn{2}{@{}l@{}}{\textbf{CLC FCE}} \\
\midrule

\textbf{Target score} & 14 \\[2pt]

\textbf{Prompt} &
You recently entered a competition and have just received this letter from the organiser. Read the letter, on which you have made some notes. Then, using all the information in your notes, write a suitable reply.

* * *

Congratulations! You have won first prize in our competition – two weeks at Camp California in the U.S.A. All accommodation and travel costs are paid for, including transport to and from the airport. We now need some further information from you:

* When would you like to travel? [only July because...]

* Accommodation at Camp California is in tents or log cabins, which would you prefer? [say which and why]

* You will have the chance to do two activities while you are at the Camp. Please choose two from the list below [Tell them!] and tell us how good you are at each one. [Tell them!]

Basketball
Singing
Swimming
Sailing
Golf
Tennis
Painting
Photography
Climbing
Surfing

Is there anything you would like to ask us? [clothes, money …?]

Yours sincerely

Helen Ryan
Competition Organiser

* * *

Write your letter.
\\
\midrule

\multicolumn{2}{@{}l@{}}{\textbf{Generated essay}} \\[3pt]

\multicolumn{2}{@{}p{0.96\textwidth}@{}}{
Dear Helen Ryan,

Thank you very much for your letter <ERR type="R" cor="and">which</ERR> congratulations for the first prize I won in your competition.

I am writing to give you the information you asked me for.

Firstly, I would like to travel in July<ERR type="MP" cor=","><MISSING/></ERR> because my school is closed during this month and I have a lot of time.

Secondly, I would prefer <ERR type="FV" cor="to stay">staying</ERR> in <ERR type="MD" cor="a"><MISSING/></ERR> tent because I think it is more exciting than <ERR type="MD" cor="a"><MISSING/></ERR> log cabin.

Thirdly, I would like to <ERR type="RV" cor="do">practise</ERR> swimming and tennis. I am very good at both of them. I <ERR type="TV" cor="have won">won</ERR> many competitions in these sports in my country and I would like to <ERR type="DV" cor="compete">competed</ERR> with other people at the Camp.

Finally, I would like to know what kind of clothes I have to bring with me and how much money I need. Is there anything else I should know?

I look forward to hearing from you soon.

Yours sincerely

Pierre Poulain
}
\\

\midrule
\multicolumn{2}{@{}l@{}}{\textbf{Write \& Improve}} \\
\midrule

\textbf{Target level} & B1 \\[2pt]

\textbf{Prompt} &
Last week, you went to a music concert. Now you are writing a letter to an English-speaking friend. Explain why you decided to go to the concert, give some information about who was performing, and describe what happened at the concert.
\\
\midrule

\multicolumn{2}{@{}l@{}}{\textbf{Generated essay}} \\[3pt]

\multicolumn{2}{@{}p{0.96\textwidth}@{}}{
Hi there! How are you? I hope you are doing great. I'm writing this letter to tell you something interesting that I did last week. Last week I <ERR type="R:VERB:TENSE" cor="went">go</ERR> to a <ERR type="R:ORTH" cor="music concert.">musicconcert.</ERR> I went with my best friend. We were so excited about it. We knew that our favorite band would come to town. We saw the band three years ago, but we didn't see them live. It was amazing. They played our favourite songs. They were all in a great mood, they were smiling all the time. My friend and I had so much fun. After the concert, we went out to eat pizza. It was delicious. See you soon! Best regards,
}
\\

\midrule
\multicolumn{2}{@{}l@{}}{\textbf{ELLIPSE}} \\
\midrule

\textbf{Target score} & 3.0 \\[2pt]

\textbf{Prompt} &
Influences of character
\\
\midrule

\multicolumn{2}{@{}l@{}}{\textbf{Generated essay}} \\[3pt]

\multicolumn{2}{@{}p{0.96\textwidth}@{}}{
Character traits are formed by influences beyond your control, or can you choose them? The character traits are formed by influences beyond your control. Because some people do things that are beyond your control, and some other people don't do anything that is beyond your control. For example <ERR type="M:PUNCT" cor=","><MISSING/></ERR>you can't change <ERR type="R:DET" cor="a">the</ERR> person <ERR type="M:NOUN:POSS" cor="'s"><MISSING/></ERR>character, and you can't influence someone to change their character. You can't control <ERR type="R:SPELL" cor="their">there</ERR> <ERR type="R:MORPH" cor="actions">act</ERR>.

Some people say that your character traits are formed by influences beyond your control. For example <ERR type="M:PUNCT" cor=","><MISSING/></ERR>you can't control <ERR type="R:OTHER" cor="yourself">your</ERR> <ERR type="R:OTHER" cor=". You">self you</ERR> just have to follow your instincts. Also <ERR type="M:PUNCT" cor=","><MISSING/></ERR>you can't control what other people think about you, and you ca<ERR type="R:CONTR" cor="n't">nt</ERR> control what other people say about you. Some people say that your character traits are formed by influences beyond your control. Because some people <ERR type="U:PRON" cor="">they</ERR> just do what they want <ERR type="M:PUNCT" cor=","><MISSING/></ERR>even if <ERR type="R:OTHER" cor="it's">its</ERR> bad for them.

Some people say that your character traits are formed by influences beyond your control. Because sometimes you do something that is good <ERR type="M:PUNCT" cor=","><MISSING/></ERR>but the next day you do something that is bad. And you can't control what your friends tell you. Because your friends tell you what to do,and sometimes your friends tell you to do bad stuff. Some people say that your character traits are formed by influences beyond your control. Because when you do something good <ERR type="M:PUNCT" cor=","><MISSING/></ERR>you feel good, but when you do something bad <ERR type="M:PUNCT" cor=","><MISSING/></ERR>you feel bad. Some people say that your character traits are formed by influences beyond your control. Because sometimes you feel like doing something bad and sometimes you feel like doing something good.

I say that your character traits are formed by influences beyond your control. Because your character traits are not really in your control, and you can't control who you are. Because some people do things that are beyond your control. Because some people do things that are not in your control. Because some people do things that are beyond your control. Some people say that your character traits are formed by influences beyond your control. Because some people do things that are beyond your control. Some people say that your character traits are formed by influences beyond your control. Because some people do things that are beyond your control.
}
\\

\bottomrule
\end{longtable}

\twocolumn

The selected CLC FCE synthetic essay follows the expected genre and addresses the prompt requirements reasonably well. It contains several plausible learner-like errors, including missing determiners, such as the omission of \textit{a} in \textit{a log cabin} and \textit{a tent}, as well as verb-form errors such as \textit{I would like to competed}. At the same time, some of the generated annotations are debatable. For example, \textit{I would like to practise swimming and tennis} is grammatically correct, yet \textit{practise} is corrected to \textit{do}. Similarly, \textit{I would prefer staying} is corrected to \textit{I would prefer to stay}, although both constructions are grammatically acceptable. These cases illustrate the occasional over-correction and annotation noise identified in the qualitative analysis.

The selected W\&I synthetic essay also follows the prompt and expected genre well. It contains plausible learner-like errors, such as the tense error in \textit{Last week I go}, where \textit{go} should be \textit{went}, as well as the orthographic error \textit{musicconcert}, corrected to \textit{music concert}. Unlike the CLC FCE example, the annotations in this essay are relatively straightforward and do not exhibit obvious over-correction. The response also contains a minor semantic inconsistency in the statement \textit{We saw the band three years ago, but we didn't see them live}, which is somewhat contradictory without further context. Overall, however, the essay remains fairly fluent and coherent, which is broadly consistent with its target B1 proficiency level. This example therefore reflects both the generally appropriate task and genre adherence and the occasional semantic inconsistencies observed in the broader qualitative analysis.

The selected ELLIPSE essay also contains several plausible learner-like errors. For example, \textit{the person character} should be written as \textit{the person's character}, while \textit{there act} appears to be intended as \textit{their actions}. Other noticeable errors include \textit{cant} instead of \textit{can't} and \textit{your self} instead of \textit{yourself}. However, the essay also exhibits a major weakness identified in the broader qualitative assessment, which is repetition. The sentence \textit{Some people say that your character traits are formed by influences beyond your control} is repeated several times, and the final paragraph is particularly repetitive, restating essentially the same idea with little additional development. The response also exhibits weak semantic development. Although it remains on topic, several supporting statements are circular or only loosely connected to the central claim. Overall, the example illustrates both the learner-like errors produced by the proposed approach and the repetition and semantic weaknesses observed particularly in the ELLIPSE samples.

Overall, the selected examples illustrate that the characteristics of the generated essays vary across corpora. While all three examples generally follow their respective prompts and contain plausible learner-like errors, they also exhibit different weaknesses, including occasional annotation noise, semantic inconsistency, repetition, and limited semantic development. These examples are intended only as qualitative illustrations of the broader patterns identified during manual inspection and should not be interpreted as representative evidence on their own. Moreover, some aspects of the assessment, particularly judgments concerning semantic coherence, repetition, and annotation quality, are inherently subjective and may reflect the authors' interpretation.

\subsection{Further Analyses of Quantitative Metrics}
In this appendix, we provide further details of the analyses presented in Section \ref{sec:analyses} that could not be included in the main paper due to manuscript length constraints.

\begin{figure*}[t]
    \centering
    \includegraphics[width=\textwidth]{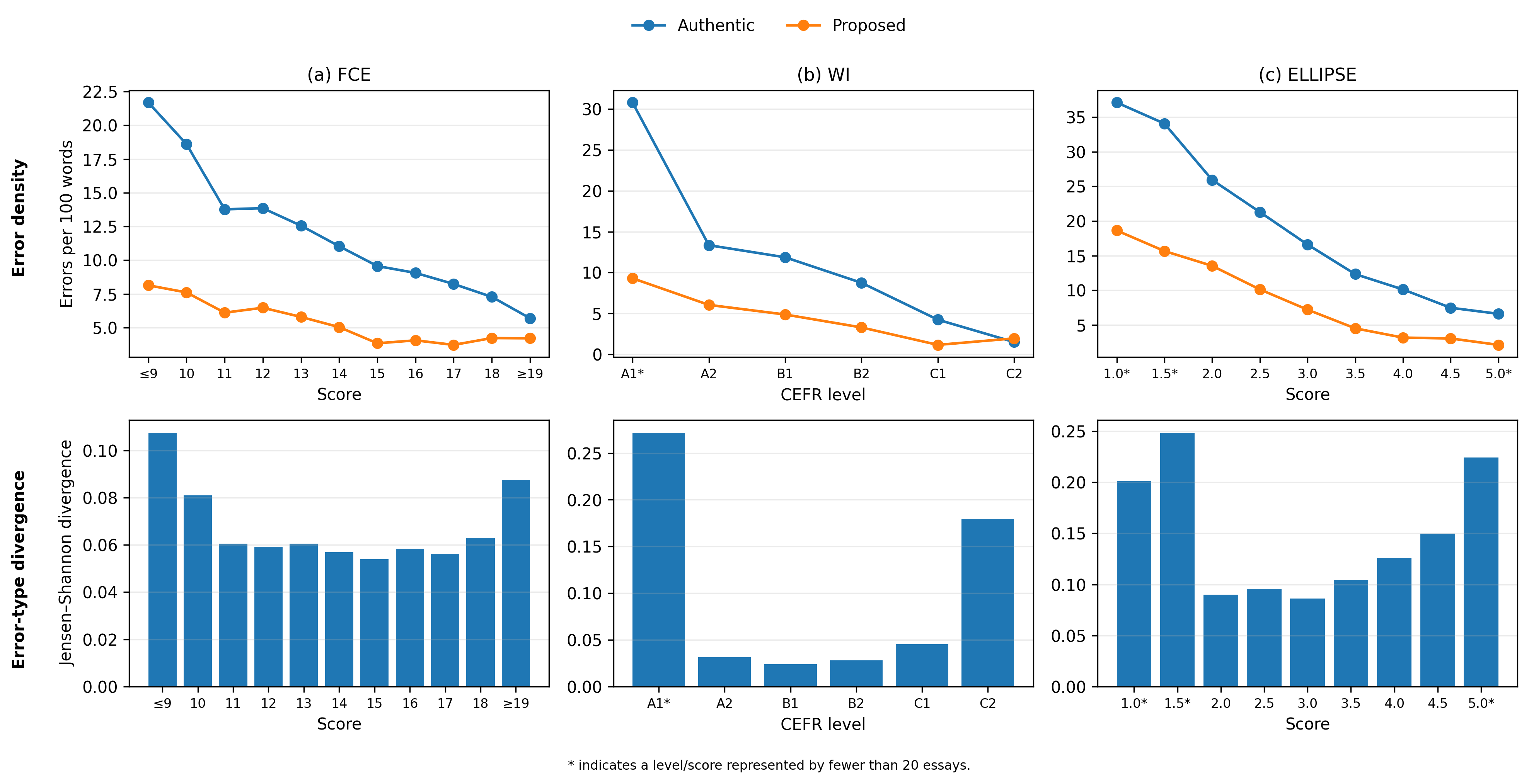}
    \caption{Comparison of error profiles between authentic and proposed synthetic training data across proficiency levels or essay scores. The top row shows error density, measured as the mean number of errors per 100 words, for authentic and proposed synthetic data. The bottom row shows the Jensen--Shannon divergence (JSD) between authentic and synthetic error-type distributions at each corresponding proficiency level or score. Lower JSD indicates greater similarity between the two distributions. Asterisks denote low-support strata.}
    \label{fig:rq2_per_level}
\end{figure*}

Figure \ref{fig:rq2_per_level} visualizes the quantitative metrics described in Section \ref{sec:analyses}. For clarity in the per-score analysis, FCE scores with sparse representation at the lower and upper ends are pooled into $\leq 9$ and $\geq 19$, respectively, to reduce instability from small sample sizes. The Jensen--Shannon Divergence (JSD; \citealp{JSD}) presented in the figure measures the divergence between two probability distributions, with lower values indicating greater similarity. It is formulated as follows:
\begin{equation} 
    \begin{aligned} 
        \mathrm{JSD}(P \,\|\, Q) &= \frac{1}{2} D_{\mathrm{KL}}\!\left(P \,\middle\|\, \frac{P+Q}{2}\right) \\ 
        &\quad + \frac{1}{2} D_{\mathrm{KL}}\!\left(Q \,\middle\|\, \frac{P+Q}{2}\right). 
    \end{aligned} \label{eq:jsd} 
\end{equation}
Here, \(P\) and \(Q\) are two probability distributions, and \(D_{\mathrm{KL}}\) denotes the Kullback--Leibler divergence \cite{kullback-leibler}. In our case, \(P\) and \(Q\) represent the error-type distributions of the authentic and synthetic essays, respectively, at each proficiency level or score. For each group, the probability of an error type is computed as its aggregated count divided by the total number of errors in that group.

To complement the JSD scores, we report the largest positive and negative differences in error-type proportions between the authentic and proposed synthetic datasets in Table~\ref{tab:error_type_differences}. This provides a more fine-grained view of which error categories are over- or underrepresented in the synthetic data. Note that the CLC FCE corpus uses the Cambridge Learner Corpus error taxonomy, whereas W\&I and ELLIPSE use ERRANT annotations. Therefore, conceptually similar error types may appear under different abbreviations in the description table. 

\begin{table*}[t]
\centering
\small
\setlength{\tabcolsep}{5pt}
\begin{tabular}{lllccc}
\toprule
Dataset & Error & Description & Auth. (\%) & Syn. (Proposed) (\%) & $\Delta$ (pp) \\
\midrule

FCE
& MP & Missing punctuation & 5.56 & 9.01 & +3.45 \\
& MD & Missing determiner & 5.37 & 8.38 & +3.01 \\
& UP & Unnecessary punctuation & 2.55 & 1.11 & -1.44 \\
& SX & Spelling confusion error & 1.37 & 0.47 & -0.90 \\

\addlinespace

W\&I
& M:DET   & Missing determiner & 5.07  & 8.51  & +3.44 \\
& R:NOUN  & Noun replacement error & 13.19 & 15.81 & +2.61 \\
& R:PREP  & Preposition replacement error & 5.42  & 3.88  & -1.53 \\
& R:SPELL & Spelling error & 6.06  & 4.77  & -1.29 \\

\addlinespace

ELLIPSE
& M:PUNCT & Missing punctuation & 14.23 & 31.12 & +16.89 \\
& R:ORTH  & Orthographic error & 6.15  & 11.72 & +5.56 \\
& R:OTHER & Other replacement error & 6.78  & 3.47  & -3.31 \\
& R:SPELL & Spelling error & 6.82  & 4.06  & -2.76 \\

\bottomrule
\end{tabular}
\caption{Largest positive and negative differences in error-type proportions between authentic and proposed synthetic data. $\Delta$ denotes synthetic minus authentic proportion in percentage points. FCE uses the native CLC FCE taxonomy, whereas W\&I and ELLIPSE use ERRANT categories.}
\label{tab:error_type_differences}
\end{table*}

Overall, the differences are relatively small for FCE and W\&I, whereas ELLIPSE shows more pronounced discrepancies, particularly for punctuation- and orthography-related errors. Across the three corpora, the largest positive differences are also consistently greater than the largest negative differences, suggesting that the most prominent distributional mismatches arise from overrepresentation of particular error types rather than severe underrepresentation. This pattern is especially apparent for ELLIPSE, where missing punctuation errors are overrepresented by 16.89 percentage points. One possible contributing factor is the source of its error supervision: unlike the other corpora, ELLIPSE does not contain native error annotations, and its tags are automatically derived using GECToR and ERRANT. Errors or systematic biases introduced by this annotation pipeline may therefore propagate into the synthetic data. However, the present analysis cannot determine whether the larger discrepancy originates from automatic annotation, characteristics of the corpus itself, or the generation process.

\section{AES Experiment Details}\label{sec:AES_details}
In this section, we provide the hyperparameters used for fine-tuning the BERT scorers as described in Section \ref{sec:experimental_setup}. Table \ref{tab:bert_hyperparameters} summarizes the hyperparameters used to fine-tune the BERT-based AES models. To focus the comparison on the downstream utility of the generated essays, all AES experiments are conducted in a prompt-unaware setting, in which only the essay text is provided to the scorer while the writing prompt is excluded. This helps ensure that the comparison reflects the usefulness of the essays themselves, rather than information provided by the prompt.

We use different training configurations for the larger-data and low-resource settings to account for the substantial difference in training-set size. In the larger-data setting, models are trained for 3 epochs with a batch size of 16. In contrast, the low-resource settings contain only 50, 100, or 200 training essays, resulting in considerably fewer parameter updates per epoch. We therefore train these models for 10 epochs with a smaller batch size of 8 to provide sufficient optimization steps while retaining the same learning rate and other training hyperparameters across settings.

For the low-resource experiments, each training size is evaluated using five sampling seeds (13, 21, 42, 87, and 100), with results reported across the resulting subsets to reduce dependence on any particular sample. Multiple sampling seeds are not applied to the larger-data setting because the complete available training set is used rather than a sampled subset. Constructing additional independent full-data settings would require changing the underlying data split; to maintain separation between the data used to train the generator and the held-out evaluation data, this would in turn require retraining the corresponding generators for each split. We therefore use a single fixed larger-data setting and reserve repeated sampling for the low-resource experiments, where sampling variability is directly relevant.
\begin{table}[t]
\centering
\small
\caption{Hyperparameters used for BERT fine-tuning in the AES experiment.}
\label{tab:bert_hyperparameters}
\begin{tabular}{ll}
\hline
\textbf{Hyperparameter} & \textbf{Value} \\
\hline
Pretrained model & \texttt{bert-base-uncased} \\
Hardware & L4 GPU \\
Maximum sequence length & 512 \\
Learning rate & $2\times10^{-5}$ \\
Training batch size & 8 \\
Evaluation batch size & 32 \\
Weight decay & 0.01 \\
Warm-up ratio & 0.0 \\
Gradient accumulation steps & 1 \\
Precision & FP16 \\
Label normalization & Min--max (regression) \\
BERT training seed & 42 \\
\hline
\multicolumn{2}{l}{\textit{Larger-data setting}} \\
Number of epochs & 3 \\
\hline
\multicolumn{2}{l}{\textit{Low-resource settings ($n=50,100,200$)}} \\
Number of epochs & 10 \\
Sampling seeds & 13, 21, 42, 87, 100 \\
\hline
\end{tabular}
\end{table}

\end{document}